\documentclass[sigconf]{acmart}

\usepackage{graphicx}

\usepackage{amsmath, amssymb, amsfonts}
\usepackage{siunitx}
\usepackage{booktabs}
\usepackage{tikz}
\usepackage{xcolor}
\usepackage{hyperref}
\usetikzlibrary{shapes.geometric, arrows.meta, positioning, calc, fit, backgrounds}
\usepackage{subcaption}
\usepackage{float}

\AtBeginDocument{%
  }

\setcopyright{acmcopyright}
\copyrightyear{2018}
\acmYear{2018}
\acmDOI{XXXXXXX.XXXXXXX}

\acmConference[Conference acronym 'XX]{Make sure to enter the correct
  conference title from your rights confirmation emai}{June 03--05,
  2018}{Woodstock, NY}
\acmPrice{15.00}
\acmISBN{978-1-4503-XXXX-X/18/06}

\begin{document}

\title{Nonholonomic Narrow Dead-End Escape with Deep Reinforcement Learning}

\author{Denghan Xiong}
\authornote{All authors contributed equally to this research.}
\affiliation{%
  \institution{ZJUI Institute, International Campus\\Zhejiang University}
  \city{Haining}
  \country{China}
}
\email{denghan.22@intl.zju.edu.cn}

\author{Yanzhe Zhao}
\authornotemark[1]
\affiliation{%
  \institution{Tianjin University}
  \city{Tianjin}
  \country{China}
}
\email{qq2954253041@tju.edu.cn}

\author{Yutong Chen}
\authornotemark[1]
\affiliation{%
  \institution{Beijing Jiaotong University}
  \city{Beijing}
  \country{China}
}
\email{23222002@bjtu.edu.cn}

\author{Zichun Wang}
\authornotemark[1]
\affiliation{%
  \institution{University of Nottingham Ningbo China}
  \city{Ningbo}
  \country{China}
}
\email{wangzichun2004@gmail.com}

\renewcommand{\shortauthors}{Xiong et al.}

\begin{abstract}
  Nonholonomic constraints restrict feasible velocities without reducing configuration-space dimension, 
  which makes collision-free geometric paths generally non-executable for car-like robots. Ackermann steering 
  further imposes curvature bounds and forbids in-place rotation, so escaping from narrow dead ends typically 
  requires tightly sequenced forward and reverse maneuvers. Classical planners that decouple global search and 
  local steering struggle in these settings because narrow passages occupy low-measure regions and nonholonomic 
  reachability shrinks the set of valid connections, which degrades sampling efficiency and increases sensitivity 
  to clearances. We study nonholonomic narrow dead-end escape for Ackermann vehicles and contribute three components. 
  First, we construct a generator that samples multi-phase forward–reverse trajectories compatible with Ackermann 
  kinematics and inflates their envelopes to synthesize families of narrow dead ends that are guaranteed to admit 
  at least one feasible escape. Second, we construct a training environment that enforces kinematic constraints and 
  train a policy using the soft actor-critic algorithm. Third, we evaluate against representative classical planners 
  that combine global search with nonholonomic steering. Across parameterized dead-end families, the learned policy 
  solves a larger fraction of instances, reduces maneuver count, and maintains comparable path length and planning 
  time while under the same sensing and control limits. We provide our project as an open source on 
  \href{https://github.com/gitagitty/cisDRL-RobotNav.git}{\textcolor{blue}{https://github.com/gitagitty/cisDRL-RobotNav.git}}

\end{abstract}


\maketitle

\section{Introduction}

Nonholonomic constraints are nonintegrable relations on configuration derivatives that restrict allowable velocities without reducing the dimension of the configuration space. As a result, an arbitrary collision-free path in configuration space is not necessarily executable by the robot. Differential geometric control links feasibility to accessibility through the Lie algebra rank condition, which establishes local controllability when the generated Lie algebra has full rank, although it does not by itself provide constructive trajectories \cite{laumond_motion_1994,laumond_guidelines_1998}.

A car-like robot with Ackermann steering is a canonical example of a nonholonomic system. Its kinematics impose a tangency constraint and a curvature bound, and the state evolves on $\mathbb{R}^2 \times S^1$ under two controls \cite{laumond_motion_1994}. In free space, the shortest feasible motions are concatenations of circular arcs and straight segments with possible reversals, that is, cusps, which formalize the need for sequences of forward and reverse maneuvers \cite{reeds_optimal_1990}. Although such systems are locally controllable, small-time local controllability fails at equilibria, so arbitrarily close configurations may still require finite-length motions with nontrivial reorientation \cite{laumond_guidelines_1998}.

Narrow environments create an additional layer of difficulty. In probabilistic roadmaps, narrow passages occupy regions of small measure in configuration space; therefore, uniform sampling captures them poorly, and specialized sampling, such as bridge tests, remains delicate \cite{zheng_sun_narrow_2005}. More broadly, sampling-based methods can require prohibitive time to discover valid connections in such regions, and the effect is amplified when nonholonomic constraints shape the reachable set \cite{orthey_section_2021}.

These limitations are particularly acute in the case of dead-end escape for Ackermann vehicles. As free space shrinks, the number of required maneuvers increases, and feasible motion often demands tightly sequenced forward and reverse actions with steering near saturation \cite{laumond_motion_1994}. Classical geometric planners that ignore nonholonomic and curvature limits cannot directly supply executable trajectories, and two-stage schemes that approximate a holonomic path and then stitch nonholonomic segments become fragile as clearances tighten near obstacles \cite{laumond_motion_1994,jiang2024hope}.

Reinforcement learning offers a complementary approach, as it optimizes sequential decisions under kinematic and dynamic constraints derived from interaction data \cite{li_survey_2025}. In parking-like tasks, learning can search for policies that reduce unnecessary gear changes while respecting safety and goal conditions. Recent studies in narrow environments show that learned policies can outperform decoupled planning and control under tight constraints when the feasible action set is explicitly structured \cite{zhengEmbodiedEscapingEndtoEnd2025a}.

We study nonholonomic narrow dead-end escape for Ackermann vehicles with three components. First, we synthesize training and test scenarios by probabilistically generating trajectories that are compatible with Ackermann kinematics, consisting of multi-phase forward and reverse segments. We then form envelopes so that each instance admits at least one feasible escape path. Second, we construct a training environment that enforces the nonholonomic constraints and train a policy using the soft actor-critic (SAC) algorithm. Third, we compare the learned policy with representative classical planners that combine global and local stages under the same scenarios and metrics.

\section{Related Work}

\subsection{Traditional Path Planning}
Nonholonomic constraints pose significant challenges to classical path planning algorithms, as a robot subject to such constraints cannot move in arbitrary directions at any given time (e.g., an Ackermann-steered vehicle cannot move sideways). Consequently, not all geometric paths generated by classical planners are physically feasible. This limitation reduces the applicability of graph-based methods such as the A* algorithm \cite{4082128} and its numerous variants \cite{zhengEmbodiedEscapingEndtoEnd2025a}, as they generally disregard kinematic constraints.

Sampling-based planners, such as Rapidly-Exploring Random Trees (RRT) and Probabilistic Roadmaps (PRM), scale well in high-dimensional configuration spaces. For instance, FastBKRRT \cite{9561207} demonstrates superior performance in motion planning for Ackermann-steered vehicles; however, its effectiveness may not extend to scenarios that demand frequent and precise pose adjustments in highly confined environments.

Obstacle avoidance strategies such as the Dynamic Window Approach (DWA), Artificial Potential Field (APF), and Timed Elastic Band (TEB) are widely employed due to their real-time responsiveness and computational efficiency. \cite{10070876} proposes a deterministic sampling method for DWA that explicitly accounts for uncertainty in differential-drive mobile robots. Additionally, reactive behaviors, such as backup-turn heuristics and velocity-space methods, can rapidly generate collision-free reversal maneuvers. Nevertheless, these approaches are typically short-sighted, prone to local minima, and limited to single-step responses.

\subsection{Path Planning with Deep Reinforcement Learning}
In contrast to traditional navigation frameworks that rely on high-precision global maps and accurate sensor inputs, DRL enables mapless navigation directly from onboard sensing, thereby enhancing adaptability \cite{9409758}. RL-based methods have been applied to nonholonomic robots in various contexts. For example, \cite{chengPathFollowingObstacleAvoidance2022} employs DDPG to balance path tracking and obstacle avoidance in wheeled mobile robots subject to nonholonomic constraints. Similarly, \cite{tai_virtual--real_2017} demonstrates virtual-to-real transfer using ADDPG with sparse lidar observations, while \cite{jengEndtoEndAutonomousNavigation2023} introduces a survival-penalty reward to address sparse feedback, showing that TD3 outperforms DDPG. Extending the application of RL to path planning, \cite{gleasonNonholonomicRobotNavigation2022} adapts DRL to Ackermann-steered vehicles navigating maze-like environments, highlighting that RL can achieve high performance even under restrictive motion constraints.

Beyond navigation, RL has also been used to resolve contention and collisions in communication systems. Shuai et al.\ integrate tabular Q-learning into framed slotted ALOHA and propose a fast-convergence MAC protocol that learns a collision-free TDMA-like schedule from local ACK feedback, achieving higher throughput and shorter convergence time than classical ALOHA variants \cite{taizhou_university_taizhou_225300_china_fast_2021}. This line of work further supports the view that RL can learn effective policies in environments dominated by collision and congestion phenomena, which is analogous to our cul-de-sac setting where nonholonomic constraints and narrow geometry create severe ``bottlenecks'' in the state space.

DRL has also been integrated with traditional path planning techniques. For instance, \cite{zhengEmbodiedEscapingEndtoEnd2025a} propose an SAC-based escape policy for differential-drive robots (robotic vacuums), augmented with A* demonstrations. Collectively, these studies demonstrate that DRL not only improves robustness over conventional path planning algorithms but also shows strong potential in addressing the challenges of navigation under stringent nonholonomic constraints.

\section{Preliminary}

\subsection{RL MDP Settings}
We cast dead-end escape as a Markov Decision Process $(S,A,P,R,\gamma)$.
The state $s\!\in\!S$ concatenates (i) a LiDAR embedding $z_{\text{lidar}}$ from a 2D 360° scan by sectorizing into $k$ bins and taking per-sector min/max after clipping at $d_{\max}$, (ii) the relative goal bearing $(d,\theta)$ from odometry/IMU, and (iii) previous control $(v,\omega)$ for action smoothing.  
The action $a\!\in\!A$ is $(v,\hat\delta)$ with bounds $|v|\!\le v_{\max}$ and $|\hat\delta|\!\le 1$.  
The reward combines sparse goal success and collision penalty with light shaping for motion/alignment:
\[
r \;=\; \lambda_{\text{col}}\,r_{\text{col}} \;+\; \lambda_{\text{goal}}\,r_{\text{goal}}
\;+\; \lambda_{\text{move}}\,(|v|+|\omega|) \;+\; \lambda_{\text{vel}}\,v\cos\theta.
\]
The objective is to learn a policy $\pi(a\,|\,s)$ maximizing expected return under these constraints.

\subsection{Robot Model (Ackermann Kinematics for Nonholonomic Platforms)}
We use a four-wheeled, front–wheel–steering platform (JetAcker) that obeys Ackermann steering geometry. Its planar kinematics are
\[
\dot{x}=v\cos\theta,\qquad
\dot{y}=v\sin\theta,\qquad
\dot{\theta}=\frac{v}{L}\tan\delta,
\]
where $L$ is the wheelbase (we also denote it by $H$ in implementation), $v$ the longitudinal velocity, and $\delta$ the front-wheel steering angle.  
Actions output a normalized steering command $\hat\delta\!\in\![-1,1]$ mapped by $\delta=\delta_{\max}\hat\delta$, hence the yaw rate used by the low-level controller is
\[
\omega \;=\; \dot{\theta} \;=\; \frac{v}{L}\tan(\delta_{\max}\hat\delta) \;\; \equiv \;\; \frac{v}{H}\tan(\delta_{\max}\hat\delta).
\]
Ackermann vehicles satisfy a nonholonomic constraint (no lateral slip).
\[
\dot{y}\cos\theta - \dot{x}\sin\theta \;=\; 0,
\]
which prohibits sideways motion and in-place rotation. The curvature is $\kappa=\tan\delta/L$ so the turning radius is $R=1/|\kappa|=L/|\tan\delta|\ge R_{\min}=L/\tan\delta_{\max}$.  
These curvature bounds and the inability to spin make cul-de-sac escape inherently multi-step (e.g., back-and-forth) and sensitive to geometry—precisely the setting where learning policies that exploit contact-free maneuvering and goal-directed bias are beneficial.

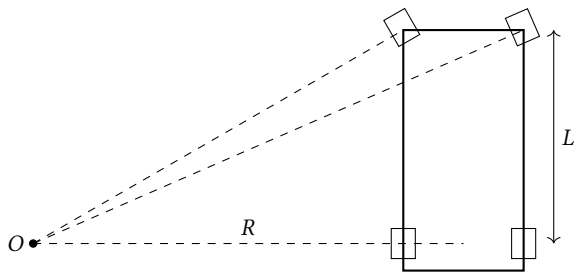
\begin{figure}[htbp]
  \centering
  \begin{tikzpicture}[scale=0.8]
    \begin{scope}[rotate=90] 
      
      \pgfmathsetmacro{\L}{3.55}
      \pgfmathsetmacro{\W}{2.0}
      \pgfmathsetmacro{\deli}{30}                          
      \pgfmathsetmacro{\R}{\L/tan(\deli) + \W/2}           
      \pgfmathsetmacro{\delo}{atan(\L/(\R+\W/2))}

      \draw[thick] (0,0) rectangle (4,2);

      \coordinate (C) at (0.45,1);
    
      \coordinate (O) at (0.45, {1+\R});
      \fill (O) circle (2pt) node[left]{$O$};

      \draw[dashed] (O) -- (C) node[midway,above]{$R$};

      \draw (0.2,2.2) rectangle (0.7,1.8); 
      \draw (0.2,-0.2) rectangle (0.7,0.2); 
      \draw[rotate around={\deli:(4,2)}] (3.8,2.2) rectangle (4.3,1.8);
      \draw[rotate around={\delo:(4,0)}] (3.8,-0.2) rectangle (4.3,0.2);

      \draw[dashed] (O) -- (4,2);
      \draw[dashed] (O) -- (4,0);

      \draw[<->] (0.45,-0.5) -- node[right]{$L$} (4,-0.5);
    \end{scope}
   \end{tikzpicture}
   \caption{\centering Schematic of Ackermann Steering Geometry}
   \label{fig:ackermann}
\end{figure}
\subsection{Problem Statement: Assumptions and Objective}
\textbf{Assumptions.} (i) Planar motion with static obstacles during an episode; (ii) onboard odometry/IMU provides bounded-drift relative pose; (iii) a 2D LiDAR with full 360° coverage; (iv) no prior map; (v) control period $\Delta t$ with bounds $|v|\!\le\!v_{\max}$, $|\delta|\!\le\!\delta_{\max}$ (equivalently $|\hat\delta|\!\le\!1$).  

\textbf{Goal.} Starting from a pose possibly inside a cul-de-sac, synthesize a collision-free control sequence that (1) reliably exits the dead-end and (2) reaches the provided goal region within horizon $T$, subject to Ackermann kinematics and curvature limits.  

\textbf{Optimization Target.} Maximize success rate and expected return while minimizing path length/steps under the above constraints.

\subsection{Environment Generation for Training and Evaluation}

We procedurally generate narrow dead-end layouts with a guaranteed feasible escape by starting from a kinematically valid seed trajectory for an Ackermann-steered robot, wrapping its exact swept area into a compact envelope, and then converting the envelope boundary into obstacles while cutting a single exit aligned with the final heading. By construction, the seed followed by a short straight extension through the exit is collision-free, which provides families of evaluation maps that are both challenging and feasible under nonholonomic constraints.

Seed trajectories are synthesized under Ackermann kinematics with bounded velocity and steering. The sampler integrates the planar model and produces multi-phase maneuver sequences that include both forward and reverse motion. Two maneuver styles are emphasized. One favors translation along a corridor and yields long, almost straight motion. The other favors tight turning with small displacements and yields turn in place like sequences without violating the no lateral slip constraint. The dataset mixes these styles in a prescribed proportion and includes instances that exit the dead end by moving forward as well as instances that exit by reversing. Each seed is densified along arclength and is then extended a short distance beyond the exit heading to define a clear goal condition and a reference demonstration.

The swept area is obtained by placing the inflated rectangular footprint at each pose of the densified seed and taking the geometric union, followed by a light smoothing and simplification step that closes microscopic gaps while preserving topology. The exit is located by marching from the final pose along the final heading until the footprint clears the envelope interior. Then, remove the boundary segment intersected by a thin strip aligned with that heading, so that a single navigable gap remains.

Two obstacle realizations are instantiated for each envelope in order to probe sensitivity to boundary density. A continuous wall variant extrudes the boundary into thin box walls of fixed thickness and height. A sparse cylinder variant samples points along the boundary at a spacing smaller than the vehicle width and places circular posts that block leakage everywhere except at the exit. Layouts are tiled across many disjoint subregions to form large batches for training and evaluation. For each instance, we export the start pose, the target region at the exit, the seed trajectory, and its corresponding control sequence with timestamps, together with simulator assets. This yields reproducible scenarios that respect nonholonomic kinematics, span long corridors, and turn dominant behaviors in controllable proportions, systematically covering both forward and reverse escape modes.

\begin{figure}[htbp]
  \centering
  \begin{subfigure}[b]{0.45\linewidth}
    \centering
    \includegraphics[width=\linewidth]{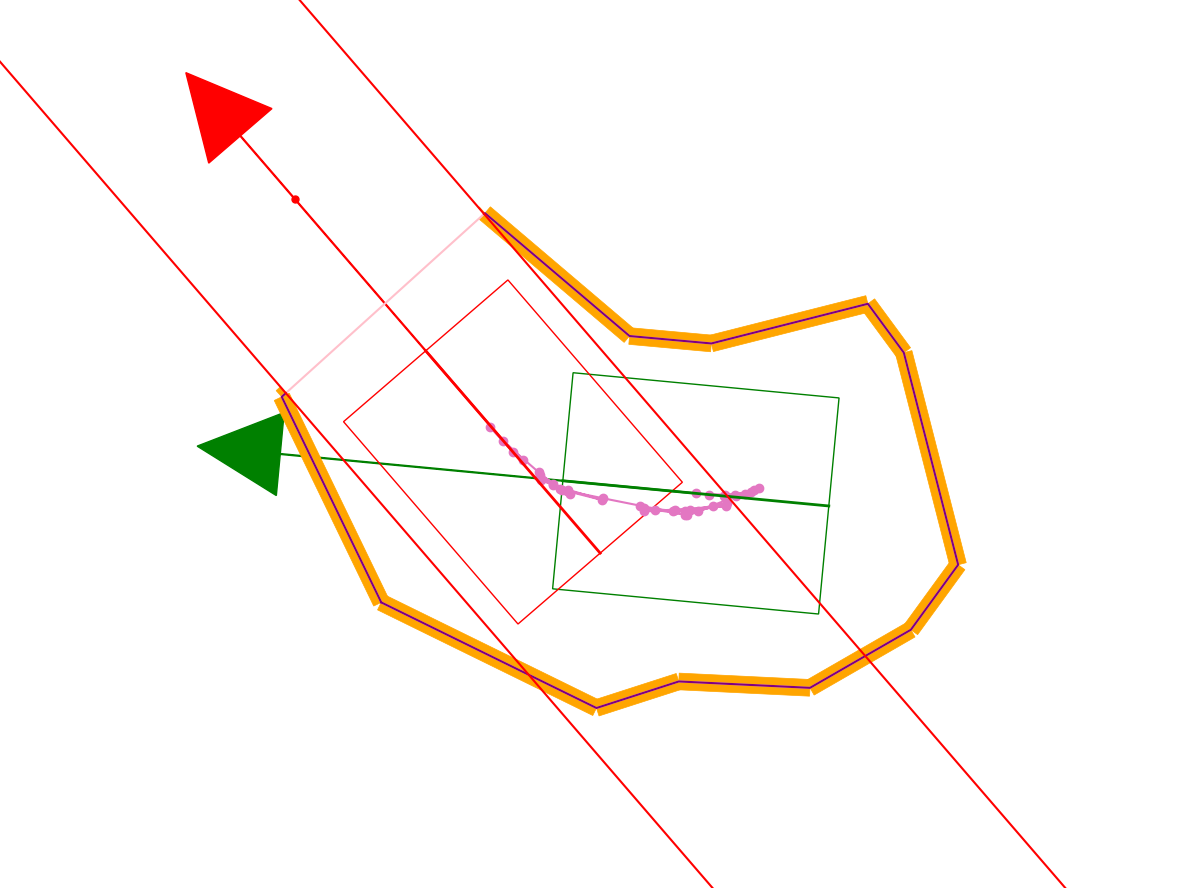}
    \caption{}
    \label{fig:subfig1}
  \end{subfigure}
  \hfill
  \begin{subfigure}[b]{0.45\linewidth}
    \centering
    \includegraphics[width=\linewidth]{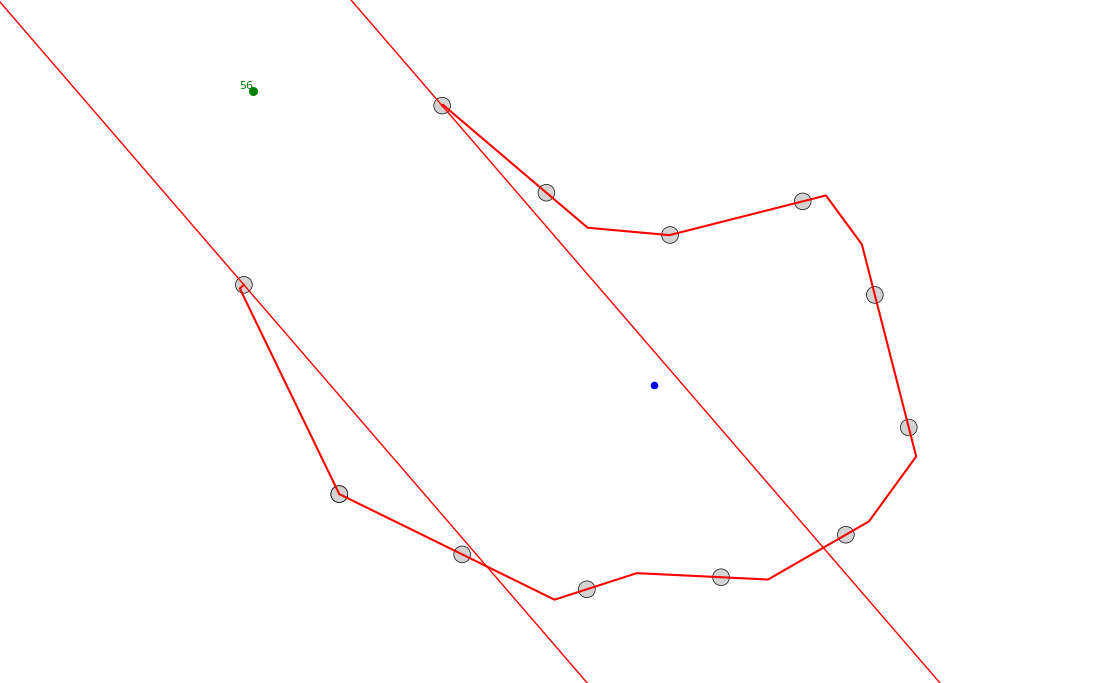}
    \caption{}
    \label{fig:subfig2}
  \end{subfigure}

  \vskip\baselineskip
  \begin{subfigure}[b]{0.45\linewidth}
    \centering
    \includegraphics[width=\linewidth]{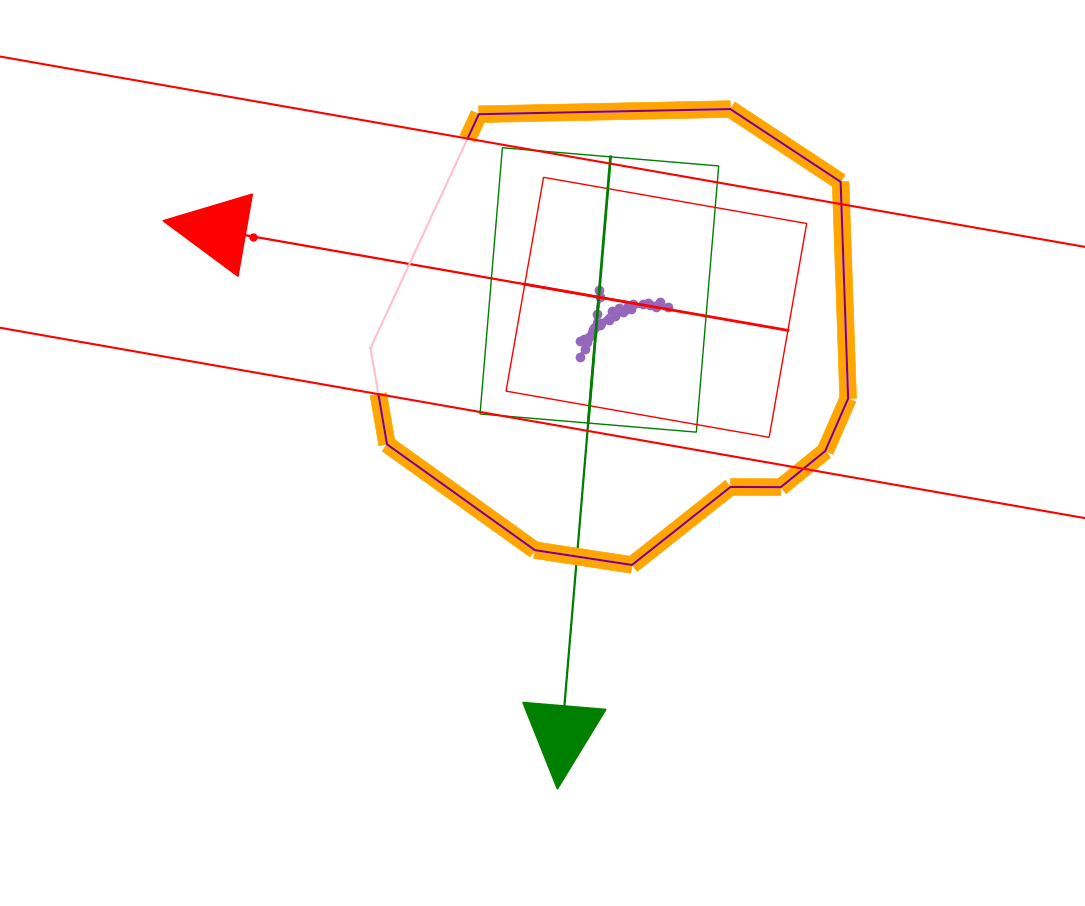}
    \caption{}
    \label{fig:subfig3}
  \end{subfigure}
  \hfill
  \begin{subfigure}[b]{0.45\linewidth}
    \centering
    \includegraphics[width=\linewidth]{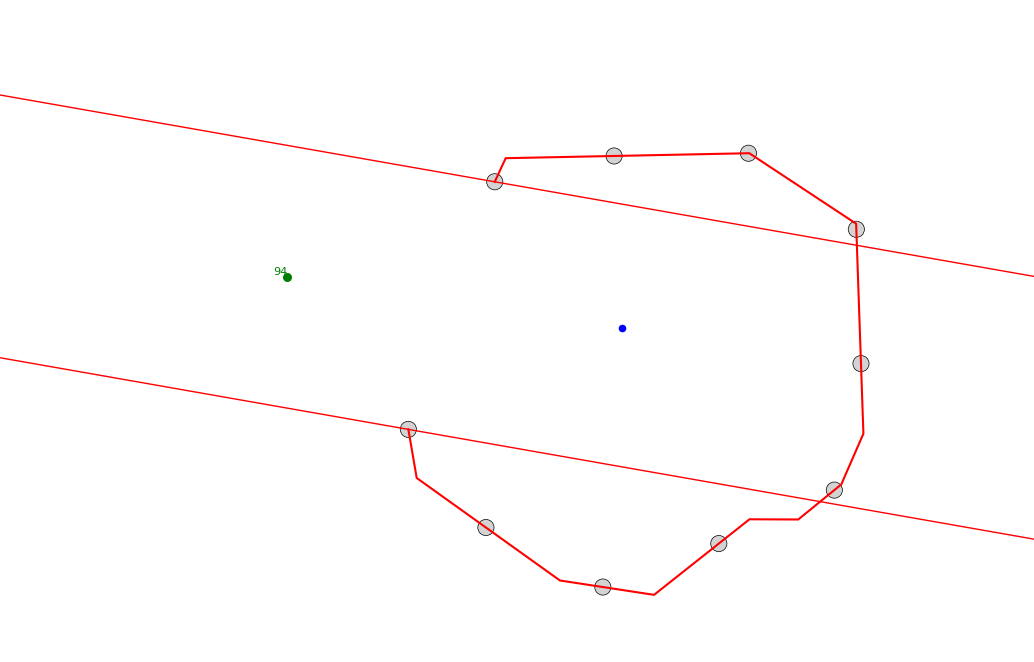}
    \caption{}
    \label{fig:subfig4}
  \end{subfigure}

  \caption{Illustration of procedurally generated narrow dead-end layouts under different trajectory styles and obstacle realizations. Subfigures (a) and (b) are generated from a trajectory biased toward long forward/reverse translation along corridors, whereas (c) and (d) are generated from a trajectory biased toward in-place maneuvering with short displacements and frequent heading changes. For each style, (a) and (c) instantiate continuous wall boundaries, while (b) and (d) instantiate sparse cylindrical boundaries. The green box and arrow indicate the start pose of the seed trajectory, and the red box denotes the terminal pose.}
  \label{fig:all_four}
\end{figure}

\section{Methodology}

\subsection{Overview}
We train a continuous-control policy for cul-de-sac escape using \emph{Soft Actor–Critic} (SAC), deployed in a ROS~2 + Gazebo loop. At each control cycle ($\Delta t=0.1$\,s), the agent reads onboard LiDAR and odometry, outputs linear speed and a normalized steering command, which are mapped to Ackermann-consistent angular velocity and executed on the robot.

\subsection{Observations and Actions}
\textbf{State.} The policy input concatenates
\[
s_t=\big[z_{\text{lidar}}^{(40)} \;\| \; d_t,\cos\theta_t,\sin\theta_t \;\| \; v_{t-1},\omega_{t-1}\big],
\]
where $z_{\text{lidar}}^{(40)}$ is the downsampled/sectorized 2D LiDAR (40 values after range clipping and \texttt{inf} masking), $(d_t,\theta_t)$ are goal bearing features from odometry/IMU, and $(v_{t-1},\omega_{t-1})$ are previous controls for action smoothing. This yields a 45-D input as in the implementation.

\textbf{Action and Ackermann mapping.}
The actor outputs $a_t=(\hat v_t,\hat\delta_t)\in[-1,1]^2$. We execute
\[
v_t=\hat v_t,\qquad
\omega_t \;=\; k_p \,\frac{v_t}{H}\,\tan\!\big(\hat\delta_t\cdot \texttt{max\_rad}\big),
\]
with $k_p{=}1$, $\texttt{max\_rad}{=}0.645$\,rad and $H{=}0.21$\,m. This realizes the Ackermann yaw-rate model $\omega{=}(v/L)\tan\delta$ (here $L{\equiv}H$) while keeping the policy output in a compact normalized range.

\subsection{Reward Design}
At each step, we combine large terminal signals (goal/collision/crash) with light shaping for movement and goal alignment:
\begin{align}
r_t =\;& 500\,\mathbf{1}[\text{goal}]
        - 100\,\mathbf{1}[\text{collision}]
        - 500\,\mathbf{1}[\text{crash}] \notag \\
     &+ |v_t|+|\omega_t|
        + v_t\cos\theta_t .
\end{align}
A goal is declared when the robot is within $0.2$\,m of the target without collision.
The terminal rewards are set on the order of a few hundred so that safety clearly
dominates: both crashes ($-500$) and minor collisions ($-100$) are far more costly
than any shaping gain. This reflects that, in our dead-end scenarios, a collision can
push the car into an unrecoverable pose and, in the real world, can damage the robot
and its sensors, so the agent is strongly discouraged from hitting walls. The small
shaping terms $|v_t|+|\omega_t|$ and $v_t\cos\theta_t$ encourage purposeful motion
and help mitigate sparse rewards: $|v_t|+|\omega_t|$ discourages freezing, while
$v_t\cos\theta_t$ rewards the component of velocity aligned with the exit direction,
effectively encouraging the robot to turn toward the opening and move along it,
whether it ultimately escapes going forward or in reverse. We assign zero reward when LiDAR data are
missing so that sensor glitches do not corrupt the learning signal.

\subsection{SAC Training Loop}
We use a standard SAC agent (stochastic Gaussian policy; twin $Q$ critics with target networks; replay buffer). Training alternates environment interaction and batched gradient updates:
\begin{itemize}
  \item \textbf{Warm start \& offline pretraining.} If enabled, we load a replay buffer from \texttt{assets/data.yml} and run $100$ pretraining iterations before online interaction.
  \item \textbf{Online interaction.} Each episode runs up to $500$ steps; transitions $(s_t,a_t,r_t,s_{t+1},d_t)$ are appended to a replay buffer of capacity $10^6$.
  \item \textbf{Updates.} Every $2$ episodes, we perform $500$ SAC update iterations with batch size $40$ (uniform sampling).
  \item \textbf{Persistence.} We periodically serialize recent trajectories back to \texttt{assets/data.yml} for future pretraining.
\end{itemize}

\subsection{Environment Sampling and Curriculum}
Episodes cycle through a configuration list (\texttt{configs.json}) that specifies the robot's start pose and target position. For each episode, we reset the simulation, place entities accordingly, and then execute the policy. Upon reaching the target, we slightly increase the internal target-distance budget (up to $8$\,m), effectively expanding the operating envelope during training while keeping the geometry fixed and reproducible.

\section{Simulation Experiments}
\begin{figure*}[t]
    \centering
    \includegraphics[width=0.3\linewidth]{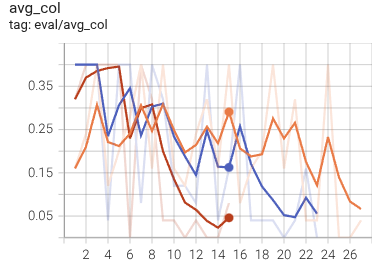}
    \includegraphics[width=0.3\linewidth]{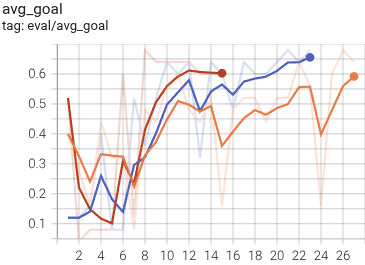}
    \includegraphics[width=0.3\linewidth]{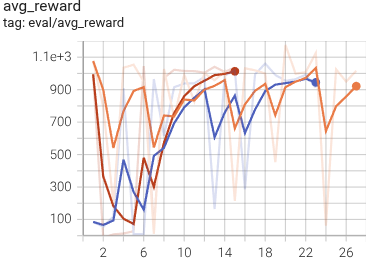}
    \caption{Training results, where the orange line represents the results on the easy environment, the blue line represents the results on the medium environment, and the red line represents the results of the hardest environment.
}
    \label{fig:results}
\end{figure*}

\subsection{Simulation Environment}
The simulation environment is established in Gazebo, where we generate an environment with a specific method: first, we define a certain position and a random direction as the initial state of a robot with a virtual car length, car width, and wheelbase width that can be defined by users and let the robot move randomly for 50 steps, then  record the final state. After that, we generate the wall around the covered area during the movement and make an exit a little wider than the car in front of the final state. Each process will generate a continuous wall version and a version with separate pillars. In an environment, there will be $n$ subsections with the process in each section, where $n$ can be defined by users. The world file is generated during the recording of the initial state and the target position for training.

\subsection{Evaluation Metrics}
\begin{itemize}
  \item \textbf{Escape success rate on unacquainted layouts with zero-shot.}  
    We evaluate the trained policy without any further learning on a held-out set of \(N\) novel dead-end maps. For each map, we run \(M\) independent episodes (with randomized start orientations) and compute
    \[
      \text{Success Rate} = \frac{\text{total escapes}}{N\times M}.
    \]
    This measures the policy’s ability to generalize directly to unseen geometries.

  \item \textbf{Average steps to escape \& collision count.}  
    Over the same test episodes, we record:
    \begin{itemize}
      \item \emph{Mean Steps to Escape}: average number of control steps taken in successful episodes.
      \item \emph{Mean Collision Count}: average number of obstacle contacts across all episodes (successful and failed).
    \end{itemize}
    These metrics quantify both efficiency (how quickly the robot escapes) and safety (how often it collides).


\end{itemize}

\subsection{Training Results}
To evaluate the proposed DRL framework, we present the training results and conduct the simulation experiment in various environments with a comparison of other algorithms. The policy was trained in Gazebo on an RTX 4070 Laptop for around 30 hours. The reward discounting factor is set to 0.99. In each epoch, we train 70 episodes.

Our training framework adopts the curriculum training method. We gradually decrease the length, width, and size of the wheelbase of the virtual car from $length = 0.47cm, width = 0.46cm, wheelbase = 0.363cm$ to $length = 0.37cm, width = 0.36cm, wheelbase = 0.263cm$ when establishing the environment. We adjust the difficulty of the environment after the training goal rate reaches $ 60\%$. Fig.~\ref{fig:results} shows the collision rate, goal rate, and reward per episode during the training process.

Although the learned policy clearly outperforms all classical baselines, it still
fails in extremely tight or highly irregular dead ends. In such cases the LiDAR
observations become nearly symmetric and the robot may commit too early to a
suboptimal turning direction, exhausting its manoeuvring room before discovering the
correct escape sequence. We also observe occasional failures when long sequences of
reversals are required, suggesting that very deep forward–reverse patterns remain
challenging for the current policy architecture and training horizon.

\subsection{Baselines and Comparative Results}

To better highlight the effectiveness of our proposed method, we compare it against three representative baselines:

\begin{table}[H]
  \centering
  \caption{Performance on unseen dead-end layouts (mean $\pm$ std (CI95)).}
  \label{tab:baseline-compare}
  \resizebox{0.48\textwidth}{!}{%
  \begin{tabular}{lccc}
    \hline
    Method & Success Rate (\%) & Steps & Collisions \\
    \hline
    \textbf{DRL (ours)   }
      & $71.85 \pm 5.68\,(3.71)$
      & $67.9 \pm 16.50\,(14.46)$
      & $5.14 \pm 3.00\,(2.63)$ \\
    Hybrid A* 
      & $7.20 \pm 0.76\,(0.67)$ 
      & $24.50 \pm 10.18\,(8.93)$ 
      & $1.30 \pm 0.60\,(0.53)$ \\
    ROS2 TEB      
      & $37.33 \pm 0.08\, (0.07)$    
      & $320.58 \pm 45.97\,(63.82)$              
      & $8.22 \pm 6.88\,(1.01)$ \\
    FTG          
      & $25.73 \pm 2.94\,(2.88)$  
      & $99.60 \pm 31.90\,(31.20)$ 
      & $12.03 \pm 0.98\,(0.96)$ \\
    \hline
  \end{tabular}}
\end{table}

Each method is evaluated over 180 dead-end instances, repeated across five random
generations (M=180, N=5, 900 trials in total). Reported means, standard deviations, and CI95
values are computed over all 900 trials.

\begin{itemize}
\item \textbf{DRL:} a plain Soft Actor–Critic agent trained end-to-end.

  \item \textbf{ROS2 TEB:} 
  the standard Nav2 global planner using the TEB local planner with a local controller
  whose parameters are adapted to our platform. TEB is widely adopted in ROS2
  navigation stacks, so it serves as a strong map-based reference for comparison in
  our setting, even though it can incur high replanning cost and get stuck in
  cul-de-sacs.

  \item \textbf{FTG (Follow-The-Gap):} 
  a LiDAR-only reactive baseline that selects the steering direction corresponding to
  the largest visible gap. We form a safety bubble around the closest obstacle,
  detect continuous gap segments, and steer toward the
  midpoint of the widest gap. To enable dead-end escape, we add a
  reversal–turn heuristic when forward clearance becomes insufficient. FTG thus acts
  as a minimal classical baseline that requires no global costmap or offline
  planning, highlighting how far a purely reactive policy can go using only
  instantaneous geometry.

\item \textbf{Hybrid A*:} 
a lattice-based kinodynamic planner used as a deterministic classical baseline.
LaserScan data are rasterised into a local 2D occupancy grid, and planning is
performed over an $(x, y, \theta)$ state lattice (0.05\,m resolution, 32 headings). 
Successor states are generated by rolling out the Ackermann kinematics under steering
limits in both directions and validated
using full-footprint collision checks. The resulting path is tracked via a
pure-pursuit controller. Hybrid A* is a
widely used planner for car-like robots, so it provides a strong kinodynamic
reference independent of learning.

\end{itemize}

We evaluate all methods on unseen procedurally generated dead-end maps, using the metrics defined previously. Table~\ref{tab:baseline-compare} summarizes the comparison.
The proposed DRL framework consistently achieves the highest escape success rate and a low collision count. 
While ROS2 TEB guarantees global path optimality with a map, it suffers from partially observable cul-de-sacs. 
Hybrid A* fails in narrow dead ends because footprint inflation and coarse lattice
resolution prune most feasible motion primitives, making multi-point turns
inexpressible. FTG fails for the opposite reason: its purely reactive gap selection cannot generate deliberate backward manoeuvres once gaps fall below the admissibility threshold. Both baselines thus struggle when clearance becomes comparable to vehicle
width.
In summary, the proposed method outperforms all classical baselines: it 
avoids the local-minima issues of TEB, overcomes the memoryless limitations 
of FTG, and achieves finer manoeuvring than Hybrid~A* without incurring 
its heavy discretisation cost.


\section{Conclusion}

We studied nonholonomic narrow dead-end escape using deep reinforcement learning and presented a pipeline that couples a feasibility-guaranteed scenario generator, a training environment that enforces car-like kinematics, and a soft actor–critic policy. On unseen layouts, the learned policy achieves higher success with fewer maneuvers and fewer contacts than representative classical planners, indicating that learning to sequence forward–reverse actions under curvature limits is effective when passages are tight and geometric connections are scarce.

Nevertheless, the overall success rate remains moderate, reflecting the inherent difficulty of navigating dead ends and escaping under nonholonomic constraints. Going forward, we will focus on improving experimental coverage and measurement to enable more informative comparisons, enhancing safety, and reducing collisions. Planned extensions include richer scenario families and ablations, unified timing and step metrics across controllers, stronger baseline configurations, and practical evaluation on real platforms with safety monitors and conservative execution to validate deployability.



\bibliographystyle{ACM-Reference-Format}
\bibliography{reference}

\end{document}